\documentclass{article}

\usepackage[preprint]{neurips_2024}

\usepackage[utf8]{inputenc} 
\usepackage[T1]{fontenc}    
\usepackage{bookmark}       
\usepackage{url}            
\usepackage{booktabs}       
\usepackage{amsfonts}       
\usepackage{nicefrac}       
\usepackage{microtype}      
\usepackage[table,xcdraw]{xcolor}
\usepackage{csquotes}
\usepackage{svg}
\usepackage{multirow}
\usepackage{tcolorbox}
\usepackage{listings}
\usepackage{adjustbox}
\usepackage{glossaries}
\usepackage{float}

\definecolor{backcolour}{rgb}{0.95,0.95,0.92}

\lstdefinestyle{mystyle}{
    backgroundcolor=\color{backcolour},
    breaklines=true
}
\title{A Frontier AI Risk Management Framework: Bridging the Gap Between Current AI Practices and Established Risk Management}

\author{%
Siméon Campos$^{1,}$\thanks{Equal contribution, corresponding authors \texttt{\{simeon,henry\}@safer-ai.org}} \quad Henry Papadatos$^{1,*}$  \\ 
\textbf{Fabien Roger$^{1,}$}\thanks{Fabien Roger was at Redwood Research while conducting this work} \quad 
\textbf{Chloé Touzet}$^1$ \quad \textbf{Otter Quarks$^1$} \quad \textbf{Malcolm Murray$^1$}\\
\\
$^1$SaferAI  \\
}

\begin{document}

\maketitle

\begin{abstract}
The recent development of powerful AI systems has highlighted the need for robust risk management frameworks in the AI industry. Although companies have begun to implement safety frameworks, current approaches often lack the systematic rigor found in other high-risk industries. This paper presents a comprehensive risk management framework for the development of frontier AI that bridges this gap by integrating established risk management principles with emerging AI-specific practices. The framework consists of four key components: (1) risk identification (through literature review, open-ended red-teaming, and risk modeling), (2) risk analysis and evaluation using quantitative metrics and clearly defined thresholds, (3) risk treatment through mitigation measures such as containment, deployment controls, and assurance processes, and (4) risk governance establishing clear organizational structures and accountability. Drawing from best practices in mature industries such as aviation or nuclear power, while accounting for AI's unique challenges, this framework provides AI developers with actionable guidelines for implementing robust risk management. The paper details how each component should be implemented throughout the life-cycle of the AI system - from planning through deployment - and emphasizes the importance and feasibility of conducting risk management work prior to the final training run to minimize the burden associated with it.
\end{abstract}

\begin{figure}[H]
    \centering
    \adjustbox{max width=0.9\linewidth,center}{%
        \includegraphics{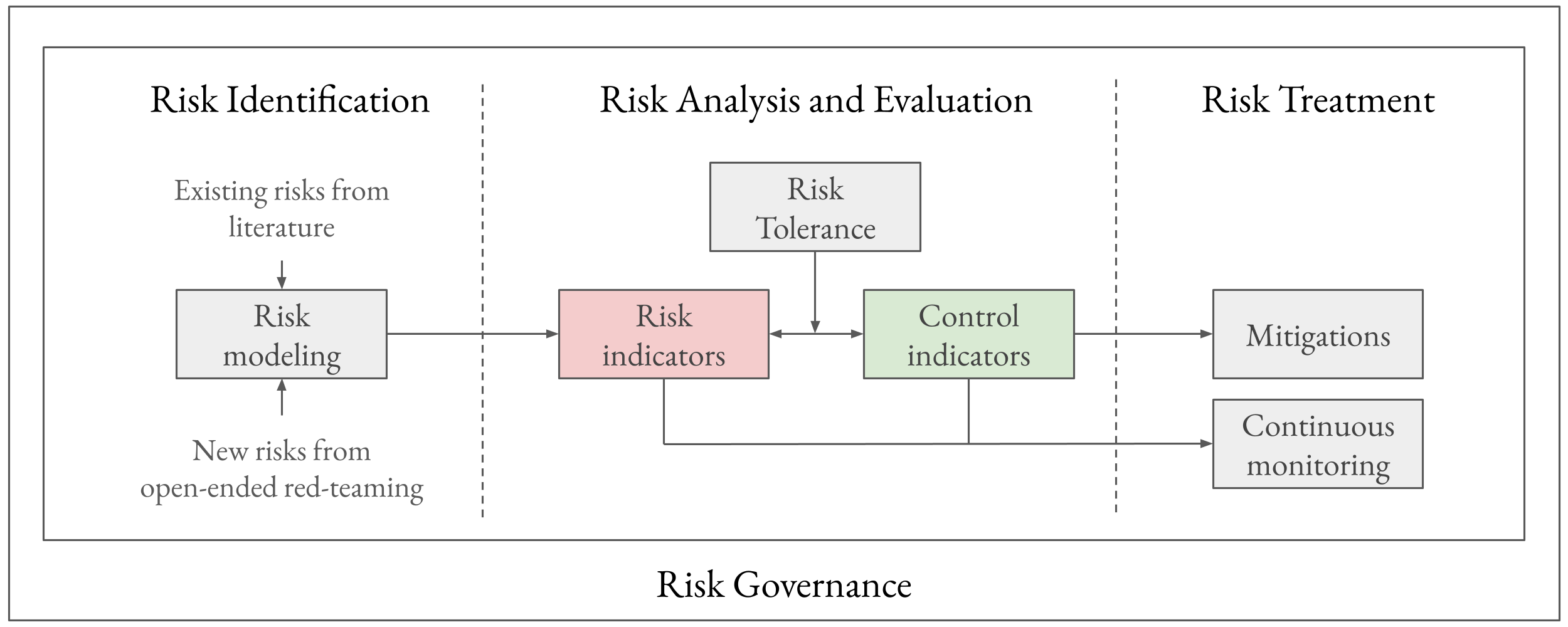}
    }
    \caption{Key components of the frontier AI risk management framework.}
    \label{fig:summary}
\end{figure}

\newpage
\section*{Executive Summary}
Frontier AI poses increasing risks to public safety and security. Managing these risks requires implementing sound risk management practices. The development of these has been the focus of several initiatives, including the Frontier Safety Commitments adopted at the May 2024 Seoul AI Safety Summit and the G7 Hiroshima Code of Conduct. This paper complements emerging practices from AI developers with risk management practices in other industries and suggestions for how to adopt them in the field of advanced AI.
 
The risk management framework introduced in this paper enables its users to implement four key risk management functions: identifying risk (\textit{risk identification}), defining acceptable risk levels and analyzing identified risks (\textit{risk analysis \& evaluation}), mitigating risk to maintain acceptable levels (\textit{risk treatment}), and ensuring that organizations have the appropriate corporate structure to execute this workflow consistently and rigorously (\textit{risk governance}). 

The \textbf{core goal of this risk management framework workflow is to ensure that risks remain below unacceptable levels at all times }through the following process: \begin{enumerate}
    \item Define a risk tolerance---a well-characterized level of risk that should not be exceeded. 
    \item Through risk modeling, operationalize this risk tolerance into pairs of empirically measurable indicators and their thresholds:
    \begin{itemize}
        \item \textit{Key Risk Indicators} (KRIs): measurable signals that serve as proxies for risks (e.g., model performance on specific tasks). KRIs should be defined with threshold values.
        \item \textit{Key Control Indicators} (KCIs): measurable signals that serve as proxies for the effectiveness of mitigations (e.g., success rate of deployment measures). KCIs should be defined with threshold values that must be met when the corresponding KRI thresholds are exceeded.  
    \end{itemize}

\item Implement mitigations to meet the required KCI thresholds whenever KRI thresholds are reached.
    
\end{enumerate}

These thresholds follow a three-way relationship: for any given risk tolerance and KRI threshold, there exists a minimum required KCI threshold that must be met to maintain risk below the tolerance.

The \textbf{risk tolerance is distinct from KRI thresholds} and can be defined in two ways: \begin{enumerate}
    \item Using quantitative probability and severity: Risk tolerance is preferably expressed as probability times severity per unit of time (e.g., less than X\% chance of \$Y million in economic damage per year).
    \item Using qualitative scenarios with quantitative probabilities: For risks where severity is difficult to quantify, risk tolerance can be expressed as a quantitative probability bound on a qualitatively defined harmful scenario. 
\end{enumerate}

\textbf{The framework takes advantage of the life-cycle of an AI system to minimize the burden on AI developers, once they complete model training}: \begin{itemize}
    \item To avoid delays during the training phase, all preparatory work that doesn't require the final model can be done ahead of time: risk modeling, defining risk tolerance, identifying KRI and KCI thresholds, and predicting required mitigations using scaling laws.
    \item This leaves only KRI measurement and open-ended red-teaming (to identify new risk factors that were not identified in the initial risk modeling) for the training and pre-deployment phase.
\end{itemize}

\textbf{On risk governance, the framework describes a corporate structure designed to ensure proportionate accounting of risks in decision-making}. It includes: \begin{itemize}
\item \textbf{Risk owner}. The risk owner is a person responsible for the management of a particular risk.
\item \textbf{Oversight}. The oversight function is board-level oversight of senior management's decision-making regarding risk.
\item \textbf{Audit}. The audit function is an independent function isolated from peer pressure dynamics that can challenge decision-making. 

\end{itemize}

\newpage

\section{Introduction}

The literature on risk management is very mature and has been refined for decades in a range of industries where safety is paramount. However, as of today, few of these principles have been applied to the risk management of frontier AI systems development, despite the mounting warnings from experts that the latter bears serious risks, ranging from enabling malicious actors to carry out cyberattacks \citep{Fang2024} or create chemical, biological, radiological and nuclear (CBRN) weapons \citep{Pannu2024}, up to potential human extinction \citep{CAIS}.

The AI industry could benefit from the lessons learned and tried and tested risk management practices in other industries. To that end, this paper presents a new AI risk management framework drawing from established risk management practices in other industries, as well as existing AI risk management approaches. This framework has four components: (1) risk identification, (2) risk analysis and evaluation, (3) risk treatment, and (4) risk governance. 

The paper starts by reviewing current AI industry practices and the existing risk management literature. It then provides an in-depth description of each of the components listed above. The paper then goes on to discuss how the framework should be implemented throughout the different phases of the life-cycle of the AI system. The final section discusses the key components and limitations of the framework. For reference, a list of abbreviations and a  glossary of technical terms are provided at the end of the paper.

\section{Background and Motivation}
\subsection{Blind Spots in Existing Safety Policies from AI Companies}

AI developers have started to propose methods to manage advanced AI risks, with a particular focus on catastrophic risks. The most prominent examples to date include Anthropic's Responsible Scaling Policy \citep{Anthropic2024}, OpenAI's Preparedness Framework \citep{OpenAI2023} and Google DeepMind's Frontier Safety Framework \citep{GoogleDeepMind2024}. At the 2024 AI Seoul Summit\footnote{The \hyperlink{https://www.gov.uk/government/topical-events/ai-seoul-summit-2024/about}{2024 AI Seoul Summit} brought together international governments, AI companies, academia, and civil society to advance global discussions on AI with a focus on AI safety.}, a further thirteen AI developers (Amazon, Cohere, G42, IBM, Inflection AI, Meta, Microsoft, Mistral AI, Naver, Samsung Electronics, Technology Innovation Institute, xAI and Zhipu.ai) agreed to also publish their safety frameworks by the time of the next AI Summit in France in February 2025 \citep{DSIT2024}.

Comparing these proto-risk management frameworks with established risk management practices, it is striking to note that these initiatives generally do not build upon or reference the risk management literature. Research analyzing these policies \citep{SaferAI2024, IAPS2024} has revealed that they deviate significantly from risk management norms, without clear justification. Several critical deficiencies have been identified: the absence of well-defined risk tolerance (\emph{risk thresholds}), the lack of quantitative or semi-quantitative assessment of risks, and the lack of systematic risk identification. The absence of a well-defined risk tolerance is especially concerning, as it can lead to risk continuously increasing, with mitigations becoming less and less adequate as capabilities increase and as it becomes increasingly hard to maintain risk below acceptable levels.

\subsection{Our Approach: Applying Risk Management Techniques to Frontier AI Development}

The field of risk management comprises a rich set of techniques in many different industries, from aviation \citep{MIT2014} to nuclear power \citep{IAEA2010}. \citeauthor{raz2005}’s (\citeyear{raz2005}) comprehensive review of existing risk management practices reveals five common steps shared in most industries:
\begin{enumerate}
    \item \textbf{Planning:} This step consists of establishing the risk context, allocating resources, setting acceptable risk thresholds, defining the governance structure and assigning roles and responsibilities.
    \item \textbf{Identification:} In this step, potential risks are identified and modeled and risk sources are established.
    \item \textbf{Analysis:} This step focuses on estimating the probability and consequences of identified risks, evaluating them, and prioritizing them.
    \item \textbf{Treatment:} At this stage, the appropriate treatment of the risks is decided and executed. 
    \item \textbf{Control \& monitoring:} Once risk treatments have been implemented, this step consists of monitoring the evolving status of identified risks and the effectiveness of treatment actions. 
\end{enumerate}

Although the concrete application of these steps to the particular context of AI risks and AI developers has not yet been studied in detail, there is some preliminary literature. \citet{Koessler2023} conduct a literature review of risk assessment techniques and analyze those that are most suitable for advanced AI developers. \citet{Barrett2023} provides a detailed and comprehensive LLM risk management profile for the NIST AI Risk Management Framework. The paper on “\emph{Emerging Processes for Frontier AI Safety}” \citep{DSIT2023} published by the UK AI Safety Institute, lists a wide range of practices to manage AI risks, including some that leverage risk management practices in other industries. A literature on safety cases and its application to AI has emerged \citep{Clymer2024} and is rapidly evolving towards increasingly concrete proposals applied to LLM cyberoffensive and misalignment risks \citep{Balesni2024,Goemans2024,Korbak2025}. 

\defcitealias{ISOIEC42001}{ISO/IEC 42001}
\defcitealias{ISOIEC23894}{ISO/IEC 23894}
\defcitealias{OECD}{OECD Due Diligence Guidance for Responsible Business Conduct}

Beyond this nascent literature, some existing standards and guidelines that harmonize risk management processes across various industries could be used for frontier AI. For example, standards such as \citetalias{ISOIEC42001} and \citetalias{ISOIEC23894}, as well as frameworks such as \citetalias{OECD} could be drawn upon.

The present framework adds to these efforts, building both on insights from traditional risk management techniques and on emerging AI-specific practices to provide a unified approach leveraging decades of risk management experience while attempting to address the unique challenges of AI development.

\section{A framework for Frontier AI Risk Management Drawing on Best Practices}

\begin{figure}
    \centering
    \adjustbox{max width=1.15\linewidth,center}{%
        \includegraphics{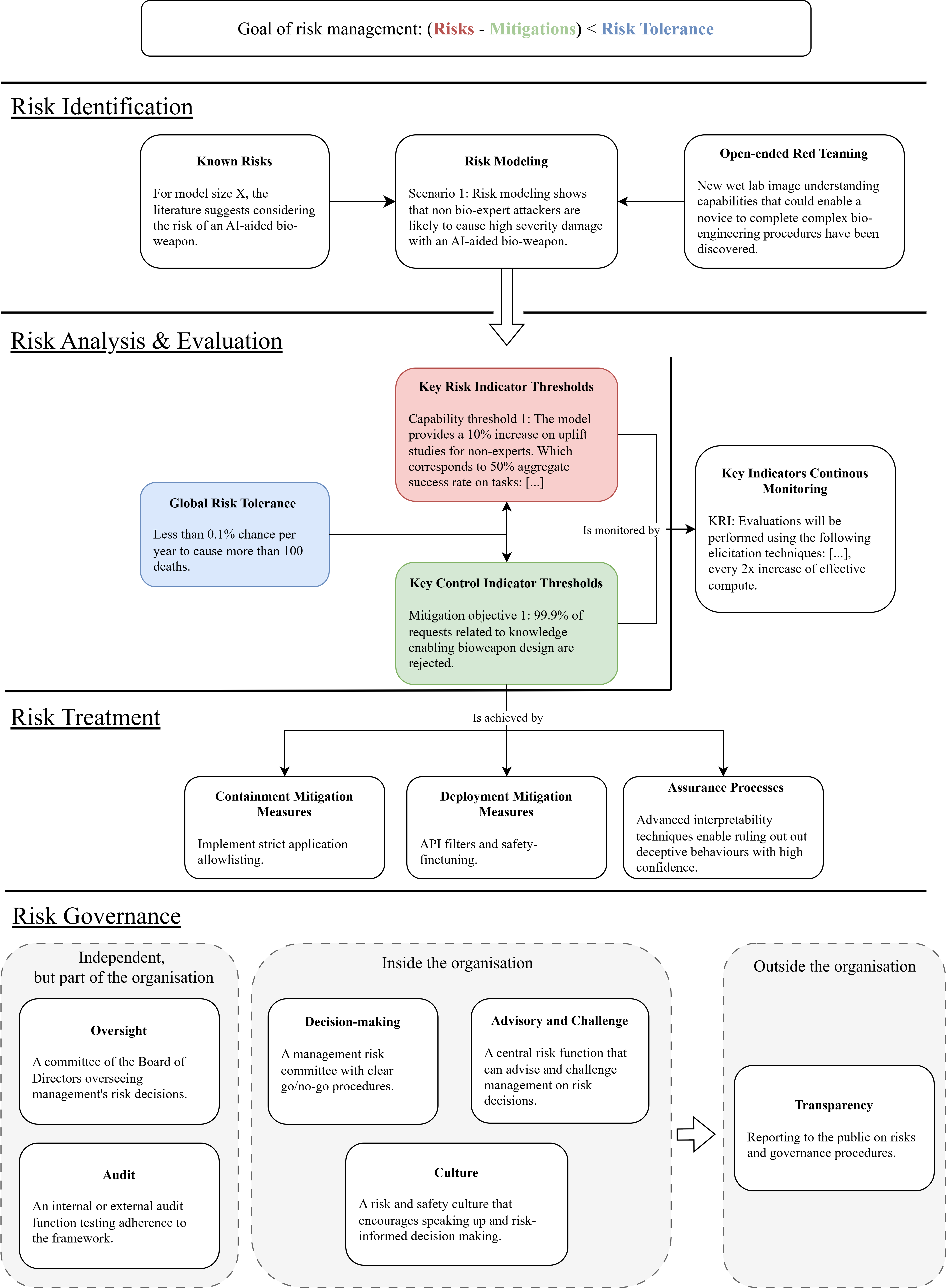}
    }
    \caption{Our complete risk management framework along with examples illustrating each component.}
    \label{fig:riskmanagement}
\end{figure}

The framework in this paper is divided into four components: risk identification, risk analysis \& evaluation, risk treatment, and risk governance, which can be compared with \citeauthor{raz2005}’s (\citeyear{raz2005}) steps described above (i.e. Planning, Identification, Analysis, Treatment, and Control \& monitoring), and the steps described in the US NIST’s AI Risk Management Framework (\emph{RMF}) (i.e. Govern, Map, Measure and Manage)  \citep{NIST2023}. Figure \ref{fig:riskmanagement} shows how the four components interact with each other and provides examples for each.
\begin{itemize}
    \item \textbf{Risk identification} is the process of identifying risks and understanding their nature (i.e. risk sources and risk scenarios)  \citep{ISO31010}. 
    \item \textbf{Risk analysis and evaluation} is a process that starts with the definition of a risk tolerance. This risk tolerance is then operationalized into risk indicators and their corresponding mitigations required to reduce risk below the risk tolerance. 
    \item \textbf{Risk treatment} corresponds to the process of determining, implementing, and evaluating appropriate risk-reducing countermeasures \citep{NISTGLOSSARY}. 
    \item \textbf{Risk governance} corresponds to the rules and procedures that structure the risk management system in terms of decision-making, responsibilities, authority, and accountability mechanisms \citep{Lundqvist2015}.
\end{itemize}

The step of risk identification aligns with Raz and Hillson’s "Identification" step and NIST’s "Map" step. Risk analysis aligns with NIST's "Measure" step and Raz and Hillson's "Analysis" step. This paper's framework also incorporates the definition of acceptable levels of risks from Raz and Hillson's "Planning" step here, as these thresholds may be specified according to the different risk domains identified in the risk identification step (e.g., economic damage thresholds for cyber risks, casualties thresholds for biological risks). 
Risk treatment corresponds to Raz and Hillson's "Treatment" and NIST's  "Manage" section. This paper also incorporates Raz and Hillson’s "Control \& Monitoring" dimension in risk treatment, because monitoring and mitigations are core to maintaining the risk below acceptable levels. Risk governance is covered by Raz and Hillson under "Planning"; however, following the NIST AI RMF, this paper argues that it should be its own component in the field of AI, given the lack of regulation and the nascency of risk management practices.

The rest of Section 3 discusses the framework's four components in detail.

\subsection{Risk Identification}
Risk identification can be divided into three aspects – classification of applicable known risks, identification of unknown risks, and modeling of the risks.

\subsubsection{Classification of Applicable Known Risks Using Taxonomies and the Literature}
Developers should address known risks in the literature using resources such as \citet{Weidinger2022} or the AI Risk Repository \citep{Slattery2024}. Developers should only exclude risks from the scope of their assessment in case of scientific agreement that the specific risk is negligible or unlikely to apply to the AI model under consideration. This decision should be clearly justified and documented.

\subsubsection{Identification of Unknown Risks Using Open-Ended Red Teaming}
In addition to risk identification based on the literature, developers should engage in extensive open-ended red teaming efforts, conducted both internally and by third parties. Open-ended red teaming aims to discover unforeseen risks and refers to red teaming practices that do not restrict exploration to predefined risk categories. The primary objectives of this red teaming effort is to identify new unforeseen risk factors that were missed in the risk modeling ahead of time. Long context length is an example: instead of models increasing biorisk through knowledge they acquired during training, models can become helpful through knowledge they acquire during inference. By modifying the risk model, it calls for a revision of the risk assessment to account for this new risk factor.

This red teaming effort must include a methodology that systematically explores the AI system for new hazards. The GPT4o model card \citep{OpenAI2024} provides a good foundation to build upon. The red team must have the appropriate expertise to properly identify hazards and should have adequate resources, time, and access to the model. Furthermore, AI developers should commit not to interfere with or suppress findings from third-party organizations, as their independent perspective is crucial to identify potential risks that may have been overlooked internally.

When novel high-severity risks are identified during open-ended red teaming, an additional risk analysis effort should be undertaken. The open-ended red teaming can be structured with several stages of triage leading up to an increasingly thorough review and risk modeling process as the vulnerabilities identified get confirmed to be significant. Early on in this triaging process, techniques such as Fishbone diagrams \citep{Coccia2018}, a structured method that helps identify and categorize the potential causes of an event, can be used. 

\subsubsection{Risk Modeling}
Risk modeling is the systematic process of analyzing how identified risks could materialize into concrete harms. This involves creating detailed step-by-step scenarios of risk pathways that can be used to estimate probabilities and inform mitigation strategies.

Established safety-critical industries have already developed sophisticated risk modeling approaches. For example, nuclear power plants and aviation use Probabilistic Risk Assessment (PRA) \citep{IAEA2010,MIT2014}, which combines event trees (showing sequential consequences of initiating events) with fault trees (working backward to identify paths leading to failures). The combination of these methods enables uncovering and modeling large parts of the risk landscape and possible paths to harm. It also allows for the creation of a range of scenarios, the likelihood and severity of which can then be estimated to assess the level of risk quantitatively.

Similar principles should be applied to AI systems. Based on risks identified in the literature and during open-ended red-teaming exercises, AI developers should create detailed scenarios mapping how an AI model's capabilities or propensities could lead to real-world harms. These scenarios should break down complex risk pathways into discrete, measurable steps. For example, one step might describe how a language model could enable criminal groups to conduct vulnerability discovery. From the full space of possible scenarios, developers should prioritize risk models that represent the most severe and probable potential harms to guide their evaluation and mitigation efforts.

The results of the risk modeling work should be well documented, including the methodologies used, the experts involved, and the list of identified scenarios. This documentation should be shared with relevant stakeholders and used to develop effective evaluation strategies, as well as effective risk mitigation strategies.

\subsection{Risk Analysis and Evaluation}
In the second step of the framework, AI developers need first to set a risk tolerance, that is, a risk level that they commit to not overshoot. Second, AI developers must operationalize their risk tolerance. This means translating the risk tolerance into concrete indicators of the level of risk---Key Risk Indicators (KRIs)---and the corresponding targets for mitigations---Key Control Indicators (KCIs)---that have to be reached.

\subsubsection{Setting a Risk Tolerance}
A risk tolerance represents the aggregate level of risk that society is willing to accept from AI systems. This tolerance can be established in different ways, depending on the field's maturity.

In the long term, risk tolerance should be expressed quantitatively as a product of the probability and severity, following practices established in other industries with societal consequences\citep{NRC1983,Nicholls2020,IAPS2024}. In the short term, given the challenges of full quantification, the severity of the risk tolerance can be expressed qualitatively using scenarios, combined with quantitative probabilities. For example: "No more than 10\% chance per year that an LLM enables an actor to damage critical infrastructure." 

In principle, establishing this risk tolerance should be the responsibility of regulators through democratically legitimate processes, as is common in other high-risk industries. For example, in aviation, it is the Federal Aviation Administration, rather than individual companies, that sets the acceptable frequency of catastrophic accidents--defined as "failure conditions that would prevent continued safe flight and landing"--to less than one occurrence per billion flight hours, equivalent to one catastrophic event per 114,155 plane years \citep{FAA1988}.

However, the AI industry currently lacks regulatory oversight and no AI developers explicitly set their risk tolerance as described above. This creates a situation where AI developers implicitly define their risk tolerance through their choice of mitigation measures and in a way that remains illegible to most third-parties. In the absence of regulation at the moment, having developers explicitly set and document their risk tolerance would enhance transparency and enable more rigorous risk management practices. When setting their risk tolerance, AI developers should still engage in public consultations and seek guidance from regulators where available. Any significant deviations from risk tolerance norms established in other industries must be justified and documented, for instance, through detailed cost-benefit analyses.

\subsubsection{Operationalizing Risk Tolerance}\label{sec:operationalization}
Risk tolerance must be operationalized into measurable criteria to be practically useful in day-to-day operations. A risk tolerance can be translated into (1) \emph{Key Risk Indicator} (KRI) thresholds, which are thresholds on measurable signals that serve as proxies for risks, and (2) \emph{Key Control Indicator} (KCI) thresholds, which are thresholds on measurable signals that serve as proxies for the level of mitigation achieved.  

These KRI and KCI thresholds must be defined in pairs following an "if-then" logic \citep{Karnofsky2024}: if a specific KRI threshold is reached, then a corresponding KCI threshold must be met to maintain risks below the risk tolerance. For any given risk tolerance level and KRI threshold, there exists a minimum required KCI threshold. This creates a three-way relationship, where setting any two parameters determines the third. In practice, frontier AI has few mitigations available to limit the most consequential risks. As a result, mitigations often tend to be a given and capabilities are the main free variables that AI developers can vary to maintain risk below the risk tolerance.

Risk models are the core element determining the relationship between KRIs, KCIs and risk tolerance. Risk models contain scenario steps, whose probability, severity or magnitude can be quantitatively estimated. This quantitative estimation could be acheived by leveraging expert elicitation techniques grounded in empirical measurements. For example, an expert Delphi survey \citep{hsu2007delphi} can help determine which cybersecurity task an LLM would have to succeed at to increase by 10 percentage points the odds that a cybercrime group manages to deploy malware in an S\&P 500 company. A set of tasks selected through that methodology, effectively a benchmark, can be used as a KRI that maps to risk tolerance. Its key difference with traditional benchmarks is that it would have a clear real-world meaning determined by experts in the way it is conceived.

If the required KCI threshold cannot be achieved, development must be put on hold until sufficient controls are implemented to meet the threshold, to ensure compliance with the risk tolerance.

\subsubsubsection{\textbf{Setting Key Risk Indicator Thresholds}}

The most important KRI threshold for AI developers will be the level of model capabilities (a primary source of risk), sometimes called capability thresholds. In addition to internal KRIs, AI developers should also identify and monitor external KRIs that signal changes in the level of risk in the external environment, such as an increased use of LLMs for cybercrime or changes in tools and scaffolding used in conjunction with the AI models.

KRI thresholds should be established for all risks identified in the risk identification phase. These thresholds should be carefully defined to provide clear signals when they are approached or breached. The precision in defining these thresholds ensures that they serve as effective triggers for implementing risk mitigation.

\subsubsubsection{\textbf{Setting Key Control Indicator Thresholds}}

For each key control indicator threshold, AI developers should define KCI thresholds, establishing a measurable target that ensures that mitigations enable risk to remain below risk tolerance even as capabilities increase. This mitigation target should build on risk modeling. By attributing probabilities to each step in the identified risk scenarios, developers can generate quantitative estimates of post-mitigation risk. These estimates must remain below the risk tolerance. 

Key Control Indicator thresholds serve two functions. The first function is to have a measurable indicator of whether mitigations are working sufficiently well. The second function is to summarize the effect of combined mitigations in a single number, making clearer how specific mitigations reduce risks. Setting KCI thresholds as an intermediary step simplifies risk modeling by allowing developers to first determine "how much" mitigation is needed before deciding "how" to achieve it through specific mitigation measures.

KCIs should be developed for three different types of mitigation:

\begin{itemize}
    \item First, \emph{containment} KCIs. To the extent possible, KCIs should be defined. If continuous metrics can't be set, they can be replaced by security levels, that correspond to ease of access to the model through theft for various stakeholders. \citep{Nevo2024}. Those should be audited by external stakeholders to be deemed valid. 
    \item Second, \emph{deployment} KCIs. These should be commensurate with the model's potential for misuse and its propensity to cause accidental risks. Such KCIs could measure how jailbreakable a model is, either during testing (e.g. a jailbreak benchmark score) or during deployment (e.g. the number of API interactions that are jailbroken per month).
    \item Third, \emph{assurance processes} KCIs. Beyond a certain level of dangerous capabilities, capability evaluations will become insufficient to demonstrate the absence of risk attached to a model \citep{Clymer2024}. At that point, models will be capable of causing harm and it will become crucial to provide affirmative evidence that they won't. We call processes that can provide such evidence assurance processes\footnote{As of early 2025, no assurance processes providing high-safety guarantees for large language models have been demonstrated yet. However, there are ongoing efforts to reach low levels of assurance \citep{greenblatt2024aicontrolimprovingsafety}.}. Assurance processes can be defined as the processes that can provide affirmative safety assurance of a model once the model has dangerous capabilities. For instance, an assurance process KCI for interpretability could be the percentage of neural network components that can be reverse engineered beyond a certain faithfulness \citep{wang2022interpretabilitywildcircuitindirect}.
   \end{itemize}

\subsubsubsection{\textbf{An Example of Risk Tolerance Operationalization}}

   Here is an illustration of how risk tolerance, KRI thresholds, and KCI thresholds may be linked:

\begin{itemize}
    \item First, the risk tolerance, when defined quantitatively, can be viewed as a risk 'budget': developers are permitted to 'spend' an overall quantity of risk when developing models, up to the defined risk tolerance. This budget can be allocated to the development of different capabilities, each associated with specific risk scenarios. For instance, developers might decide to focus on improving coding capabilities, which is associated with scenarios of AI-enabled cyberattacks. In this case, they might allocate a larger portion of their risk budget to address potential cyber-misuse risks. The allocation of risk budgets should be guided by expert input, considering the expected benefits of each capability, strategic priorities, and the intended focus of the model. 
    \item Second, KRI and KCI thresholds are determined using expert inputs and in-depth risk modeling. As an illustrative fictional example, "\emph{if a model reaches 60\% on Cybench \citep{zhang2024cybenchframeworkevaluatingcybersecurity} (KRI threshold), then maintaining cyber security level 3 \citep{Nevo2024} (KCI threshold) is required to keep the risk of causing economic damages below \$500M below 1\% per year.}" 
    \footnote{Establishing such quantitative links between capabilities and risks using current AI risk management practices remains challenging. Further advances in quantitative AI risk assessment methods will be necessary to enable proper risk management.}

\end{itemize}

\subsection{Risk Treatment}
The third component of the risk management framework is risk treatment, where mitigation measures are implemented to control the level of risk within the limits established in the earlier step (i.e., reach the KCI threshold and remain below the risk tolerance). 

\subsubsection{Implementing Mitigation Measures}

AI developers should operationalize--at least internally--their KCI thresholds, defined in the risk analysis and evaluation step (Section \ref{sec:operationalization}), into mitigation measures. 

Containment measures are largely information security measures that allow controlling access to the model for various stakeholders. For the potential loss of control risks, containment also includes containing an agentic AI model. Examples include extreme isolation of weight storage, strict application allow-listing, and advanced insider threat programs \citep{Nevo2024}.

Deployment measures aim to mitigate risks resulting from the usage of the model. These are the set of measures that allow controlling the potential for misuse of the model in dangerous domains and its propensity to cause accidental risks. Examples include API input / output filters, safety fine-tuning, and knowing your customer policies \citep{DSIT2023}.

Finally, as discussed above, past a certain level of dangerous capability, implementing credible mitigation measures is likely to require setting assurance processes, supported by evidence that these are enough to achieve the risk tolerance. Understanding that those assurance processes don't yet exist, AI developers should have credible plans towards the development of such processes. For their assurance process plan, AI developers should clearly specify the underlying assumptions that are essential for its effective implementation and success. Examples include using advanced interpretability to reliably detect deception and formal verification \citep{dalrymple2024guaranteedsafeaiframework}.

\subsubsection{Continuous Monitoring and Comparing Results with Pre-determined Thresholds}

Unlike in some other industries where risks primarily materialize when the final system is deployed (e.g., an aircraft's safety risks emerge once it starts flying), AI systems can pose risks throughout their development cycle. For instance, loss of control scenarios could materialize during the training process itself, requiring continuous monitoring and risk mitigations well before deployment. 

This means that capability evaluation is not a one-off affair, but should be repeated regularly during training and during deployment. Developers must therefore implement continuous monitoring of both KRIs and KCIs to ensure that KCI thresholds are met once KRI thresholds are crossed according to the predefined "if-then" statements established in the risk analysis and evaluation phase.

AI developers should establish rigorous evaluation protocols designed to produce upper bound estimations of AI systems' capabilities in order to ensure that KRI thresholds are not crossed unnoticed. These protocols should specify the evaluation frequency in terms of both the relative variation of effective computing power used in training and fixed time intervals to account for post-training enhancements \citep{Anthropic2024}.\footnote{Post-training enhancements refer to modifications that can increase model capabilities after the training is complete, such as prompt engineering, scaffolding, or the integration of external tools. These enhancements can significantly alter the capabilities of a model without requiring additional training} Evaluations must be performed sufficiently frequently. The elicitation methods used during the evaluations must be comprehensive enough to match the elicitation efforts of potential threat actors. Increased test-time computing power must be included in elicitation efforts. The evaluation environment and methodology must be documented, including specifying how post-training enhancements are factored into capability assessments.

Similarly, AI developers should monitor KCIs to ensure that mitigation measures are functioning appropriately and are meeting the KCI thresholds.

Independent third parties should vet evaluation protocols. These third parties should also be granted permission and resources to independently perform their evaluations, verifying the accuracy of the results. In addition, AI developers must commit to sharing the evaluation results with relevant stakeholders as appropriate.

\subsection{Risk Governance}
The last part of the risk management framework is risk governance. Risk governance consists of defining the decision-making structure for the risk identification, analysis and evaluation, and treatment components. In essence, it consists of defining "who does what" and "who verifies how it is done," ensuring there are clear roles and responsibilities for decision-making in the risk management processes. 

Risk governance can be seen as a set of interlocking components that play unique roles and jointly form a cohesive governance structure. As seen in Figure \ref{fig:riskmanagement} above, these components can be placed in six distinct categories, each fulfilling different purposes. Three categories relate to risk decisions inside the organization, two categories to oversight that is inside the organization, but independent, and one category to the communication of decisions outside the organization. The remainder of this section outlines the six categories, how they relate to each other, and why each one is essential.  

\subsubsection{Decision-Making}
The first category of risk governance components is at the core of risk management, that is, the decisions made by senior management that create or mitigate risk. Best practices from other industries include the establishment of clear risk ownership, with designated senior managers responsible for specific risks and a dedicated senior-level committee for risk-focused decision-making \citep{COSO2017}. Senior management should have clear go/no-go decision protocols and rules to follow in their decision-making \citep{Eisenhardt1989,Hammond1998}. Finally, since AI risk evolves rapidly, there should also be a clear escalation process for rapid decision-making when there are changes in risk levels\citep{COSO2017}.

\subsubsection{Advisory and Challenge}
The second category consists of risk experts who advise and challenge senior management on their decisions. The senior managers making risk decisions need to be distinct from those advising on the decisions, to avoid conflicts of interest. Therefore, there should be a senior executive responsible for risk management processes (often called a Chief Risk Officer) who is accountable for the risk management processes, but is importantly not a risk owner making risk decisions themselves. Without them, senior management's decisions are likely to be subject to short-term pressures such as deadlines or performance, at the expense of safety. 

To provide support to the Chief Risk Officer, it is common in many industries to have a central risk function. This function is in charge of the risk management process, providing support and advice, challenging management on the soundness of their decision-making, and tracking and monitoring risks. In most industries, this function is known as Enterprise Risk Management (ERM). This function also prepares appropriate risk information for senior management and the Board.

\subsubsection{Culture}
The third category consists of the components that influence everyone in the organization, through the culture. The major decisions on risk will be made by senior management, advised by risk experts. However, risk is created, influenced and observed by people who are several levels below senior management. This creates the need for a focus on risk culture (known in some industries as safety culture). This is the "set of norms, attitudes and behaviors related to awareness, management and controls of risks"  \citep{ECB2023}. These shape day-to-day decisions impacting risks.

Another key element of culture is speak-up culture, “a workplace environment where employees feel comfortable speaking their minds, sharing their ideas, and raising concerns without fear of negative consequences” \citep{West}. In aviation, this is often called a “just culture”, a culture without retaliation for speaking up and reporting problems \citep{Parker}. A key feature of a speak-up culture is whistleblowing - processes for anonymously reporting issues. This is important at AI developers, where new risks and safety issues might be discovered serendipitously.

All aspects of culture are ultimately driven by the “tone at the top”. This can be defined as “top management’s way to express [...] values pursued in the organization and provide guidance to employees” \citep{EweltKnauer2020}. It refers to the communications made by senior leadership on risk and safety and is a core cultural element influencing how the whole organization will take decisions that impact risks.

Finally, incentives to perform and report positive information at each level of management can filter out negative information and cause senior management to receive only a small fraction of the information relevant to risk decision-making. A poor organization-wide culture can lead the leadership to systematically underestimate risk.

\subsubsection{Oversight}
Board-level oversight is necessary to provide checks and balances to senior management. The Board of Directors is the only body that provides oversight over the decisions of senior management and is specifically tasked with considering the longer time horizon. Without the Board, senior management can overly optimize for short-term performance, ignoring risks. The Board typically has one subcommittee focused on risk, either called an Audit committee or a Risk committee. Audit committees are one of the standard committees, alongside Remuneration and Nomination, that exist in most publicly listed companies. They are prescribed in US listed companies under the Sarbanes-Oxley regulation of 2002 \citep{Coates2007}. In the Financial Services industry, there are also often dedicated risk committees. There can also be governing bodies separate from the Board that fulfill a similar function, such as Anthropic's Long-Term Benefit Trust (LTBT) \citep{Anthropic2023}.

\subsubsection{Audit}
The fifth category consists of independent groups that test the efficacy and sufficiency of the risk management framework and of the risk mitigations. Without independent groups providing regular checks on risk management processes, they can quickly deteriorate in quality and become purely performative. Audits are provided by internal auditors and/or external auditors. In both cases, they are independent from peer pressure dynamics occurring within the teams dealing with the risk. Internal Audit is a part of the organization, but has a unique reporting line to the Audit committee of the Board to avoid conflicts of interest with the business. It has the mandate to investigate any process in the organization. This is a common function in publicly listed organizations and is a listing requirement of many stock exchanges, such as the New York Stock Exchange \citep{NYSE}.

Internal audit teams may lack expertise in certain risk areas, particularly technical risks. Therefore, most industries also use external auditors. These are specialists from outside the organization who are brought in to provide additional assurance expertise. Publicly listed companies in the United States for example must, per the SEC's (Securities and Exchange Commission) regulations, employ an external auditor for their financial statements \citep{SECauditors}. External audit providers can also provide assurance in other areas, such as quality or cybersecurity. In AI, external auditors are already used for red-teaming and model capability evaluations.

\subsubsection{Transparency}
The last category of risk governance focuses on external communication of risks and decision-making. Without transparency, the public and regulators would have no way of judging whether the company adequately manages risk.

The first type of communication is external disclosure of the risks faced by the organization. In the United States, this is required for listed companies and is provided in the annual report\citep{WhiteCase2023}. In the case of AI, given the broad nature of risks the technology poses, this should be broadened to include risks to society from the company's products.

It is also important that the organization discloses details on their governance structure to provide transparency on how risk is managed. In other industries, annual reports must disclose elements of governance, both in the United States \citep{SECexchange} and in other jurisdictions \citep{UNCTAD2006}.

In the case of AI, with rapidly changing risks and significant potential for hidden errors, it is also vital that organizations provide external incident reporting. This type of reporting can be addressed to industry bodies, such as the Frontier Model Forum \citep{FMF2023}, or to regulators. 

\section{Implementing the Frontier AI Risk Management Framework}
\subsection{Risk register}
Throughout the risk management process, AI developers should maintain a continuously up-to-date risk register, which serves as the central repository for documenting and tracking all identified risks and their associated mitigation measures. This is key as an internal tool for all employees of the company to have up-to-date information regarding the status of each risk. 

For each risk identified, we suggest that the risk register could document the following information:

\begin{itemize}
    \item \textbf{Risk owner:} The individual within the organization who is responsible for ensuring the appropriate management and mitigation of risk. 
    \item \textbf{Risk level:} The inherent risk level (pre-mitigation) and the residual risk level (post-mitigation) of the riskiest model.
    \item \textbf{Key Risk Indicators (KRIs):} Proxy measures that are tracked to monitor and assess risk status (for more definition, see Section \ref{sec:operationalization}).
    \item \textbf{Mitigation status and Key Control Indicators (KCIs):} The Key Control Indicators and measures, together with their effectiveness in reducing risk level (for a more detailed definition, see Section \ref{sec:operationalization}).
    \item \textbf{KRI / KCI mapping:} The description of levels of KRIs at which particular KCIs must be achieved, along with the corresponding expected residual risk.
    \item \textbf{Action plan:} A detailed plan that outlines the key actions required to manage risk in the short and medium term, including timelines, responsibilities, and allocation of resources.
\end{itemize}

\subsection{Life-cycle}

In order to implement the risk management framework described in this paper, it is necessary to understand at what point of the AI life-cycle components should be executed. This paper argues that conducting the majority of risk management activities during the planning phase, before the final training run begins, appears adequate for two key reasons. First, the planning phase provides the necessary time horizon, as many crucial elements, such as implementing robust containment measures, require a long time. Second, conducting risk management activities in parallel with capability research and development, well before deployment, helps avoid the commercial pressure to compromise on safety measures to meet release deadlines. This section examines the appropriate timing for each framework component in the planning, training, and post-deployment phases.

\subsubsection{Planning Phase}

In the planning phase, AI developers allocate resources for model training, determine model parameters such as compute requirements, and conduct preliminary experiments before the final training run begins. A large part of the risk management effort can already occur in this phase. 

During planning, developers should notably perform risk identification based on known risks from the literature and perform the corresponding risk modeling. The risk tolerance should be established, along with defined KRI thresholds and their corresponding KCI thresholds that are sufficient to maintain risks below the tolerance level. Using scaling laws \citep{ruan2024observationalscalinglawspredictability,kaplan2020scalinglawsneurallanguage, hoffmann2022trainingcomputeoptimallargelanguage}, developers can anticipate which KRI thresholds are likely to be crossed during development and can therefore implement appropriate mitigation measures ahead of time to ensure that they will be ready when needed. However, even though scaling laws provide valuable predictive power for capability trends, empirical evaluation remains essential to accurately measure KRIs throughout the training phase and verify these predictions.

\subsubsection{Training Phase}

During the training phase, AI developers should conduct open-ended red teaming to identify any unexpected risks or emerging capabilities. If significant findings emerge from this analysis, developers should update their risk models, KRI and KCI thresholds as needed.
Throughout the training, developers must continuously monitor KRIs. When KRI thresholds are crossed, they should ensure that corresponding mitigation measures are implemented to meet the KCI thresholds. Parallel continuous monitoring of KCIs is also essential to verify that these thresholds continue to be met.

\subsubsection{Post-Deployment Phase}

After deployment, developers must maintain continuous monitoring of both KRIs and KCIs to ensure all thresholds remain within acceptable bounds. This monitoring should include vigilance for improvements in post-training enhancements that may affect the level of risk that a model induces. Additionally, deployment could enable the measurement of KRIs more directly correlated with real-world harms, such as tracking real-world incidents using API logs. Some KRI thresholds could also be directly defined for deployment, in case the risk modeling done ahead of deployment was overly optimistic. For instance, real-world cybersecurity incidents caused by LLMs crossing significant thresholds could induce new mitigations.

\section{Conclusion}

This paper presents a risk management framework for frontier AI developers. This framework, grounded in established risk management practices in other industries and existing AI risk management approaches, comprises four dimensions:

\begin{enumerate}
    \item Risk identification 
    \item Risk analysis and evaluation
    \item Risk treatment
    \item Risk governance
\end{enumerate}

One of the key elements missing from current AI risk management practices and mandated by this framework is the explicit setting of a quantitative risk tolerance. Although established high-risk industries such as aviation or nuclear power have clear regulatory frameworks defining acceptable risk levels, the AI industry currently lacks such guidance. This creates a situation where AI developers implicitly define their risk tolerance through mitigation choices rather than explicitly. Until regulatory frameworks emerge, developers must proactively set and document their risk tolerance. This represents a pragmatic intermediary step to enable rigorous risk management.

Although this framework provides a structured approach to AI risk management, several methodological gaps must be addressed for its full implementation. The primary challenges lie in risk modeling and quantitative assessment. The field of frontier AI still lacks a detailed understanding of how harms could materialize in the real-world, while this knowledge is crucial to define appropriate KRI and KCI thresholds. Furthermore, current quantitative risk assessment methods are currently insufficient to rigorously demonstrate that KCI thresholds maintain risks below the risk tolerance. On the mitigation side, research is needed to ensure mitigation measures scale with advancing AI capabilities, particularly in developing assurance processes that can provide meaningful safety guarantees for increasingly powerful systems.

This work aims to encourage AI developers to adopt and develop more rigorous and standardized risk management processes by providing a structured approach to AI risk management practices. As the field of AI development continues to advance, this paper contributes to AI risk management practices keeping pace with the evolution of AI capabilities, so that society as a whole can both reap the benefits of advanced AI and avoid the harms. 

\pagebreak

\section*{Acknowledgments}
We thank the following individuals (listed alphabetically) for their valuable input on this work and its precursors: Amin Oueslati, Anthony Barrett, Buck Shlegeris, Chris Painter, James Gealy, Jide Alaga, Marius Hobbhahn, Patricia Paskov, Stephen Casper, Zach Stein-Perlman.

All comments were made in personal capacity and do not represent endorsement of the views expressed in this paper. Mistakes are our own.

\section*{Abbreviations}
\begin{itemize}
    \item \textbf{AI} - Artificial Intelligence
    \item \textbf{API} - Application Programming Interface
    \item \textbf{CBRN} - Chemical, Biological, Radiological and Nuclear
    \item \textbf{ERM} - Enterprise Risk Management
    \item \textbf{IEC} - International Electrotechnical Commission
    \item \textbf{ISO} - International Organization for Standardization
    \item \textbf{KCI} - Key Control Indicator
    \item \textbf{KRI} - Key Risk Indicator
    \item \textbf{LLM} - Large Language Model
    \item \textbf{LTBT} - Long Term Benefit Trust
    \item \textbf{NIST} - National Institute of Standards and Technology
    \item \textbf{NYSE} - New York Stock Exchange
    \item \textbf{OECD} - Organisation for Economic Co-operation and Development
    \item \textbf{PRA} - Probabilistic Risk Assessment
    \item \textbf{SEC} - Securities and Exchange Commission
    
\end{itemize}

\section*{Glossary}
\begin{itemize}
    \item \textbf{Assurance processes:} Processes that can provide affirmative safety assurance of an AI model once the model has dangerous capabilities.
    \item \textbf{Audit:} Process by which independent (internal or external) evaluations are conducted to verify the effectiveness, accuracy, and compliance of the risk management framework and its measures.
    \item \textbf{CBRN Weapons:} Chemical, Biological, Radiological and Nuclear Weapons. In the context of AI risk management, used to discuss the potential for AI systems to be misused in the development of high consequence weapons.
    \item \textbf{Capabilities thresholds:} Defined levels of an AI system's performance or capabilities that, when reached, require implementation of specific mitigation measures to prevent risk exceeding established risk tolerance.
    \item \textbf{Containment measures:} Mitigation strategies focused on controlling access to the AI system.
    \item \textbf{Deployment measures:} Risk mitigations that allow controlling the potential for misuse of the model in dangerous domains and its propensity to cause accidental risks.
    \item \textbf{Enterprise risk management:} An organizational framework for identifying, assessing, and managing risks across the entire enterprise.
    \item \textbf{Frontier AI:} Advancements, innovations and models that push the limits of current AI technology and capabilities.
    \item \textbf{Key Control Indicator (KCI):} Measurable targets representing the effectiveness of mitigation measures.
    \item \textbf{Key Risk Indicator (KRI):} Measurable signals that act as proxies for risk. KRIs help monitor risk levels in the system and serve as triggers for when additional mitigations should be applied.
    \item \textbf{Open-ended red teaming:} A form of red teaming that aims at discovering unforeseen risk or risk factors, by not restricting  exploration to predefined risks.
    \item \textbf{Red teaming:} A practice where experts challenge and probe an AI system to identify vulnerabilities and potential risks, designed to mimic adversarial or unforeseen conditions.
    \item \textbf{Risk analysis and evaluation:} A phase where risks are assessed by setting a risk tolerance and translating it into measurable indicators. This process involves determining the probability and severity of risks and prioritizing them for further action.
    \item \textbf{Risk governance:} A system of rules, processes and practices that define how an organization makes decisions regarding risk management. It covers the allocation of responsibilities, decision rights, oversight mechanisms, and procedures for external transparency and reporting.
    \item \textbf{Risk identification:} The process of recognizing and categorizing potential hazards, risk sources and scenarios, including both known and unknown risks.
    \item \textbf{Risk modeling:} A process of constructing detailed, step by step scenarios that describe how identified risks might materialize into real-world harm.
    \item \textbf{Risk register:} A central, continuously updated document that tracks all identified risks, including information such as risk owners, risk levels, associated KRIs and KCIs, and action plans for mitigations.
    \item \textbf{Risk tolerance:} The aggregate level of risk that society or AI developers is willing to accept.
    \item \textbf{Scaling laws:} Empirical relationships predicting how changes in computing resources or model size affect an AI system's performance. In the context of risk management, these help forecast when a model might reach critical thresholds that necessitate mitigations.
    \item \textbf{Speak-up culture:} Also known as "just culture", an environment in which employees feel empowered and safe to report risks, concerns or failures without fear of retaliation.
    \item \textbf{Tone at the top:} The ethical climate established by an organization's senior leadership.
\end{itemize}

\newpage

\bibliographystyle{chicago} 
\bibliography{references}

\end{document}